\documentclass[sigconf]{aamas} 

\usepackage{balance} 
\usepackage{hyperref}
\usepackage{caption}
\usepackage{subcaption}
\usepackage[commandnameprefix=always]{changes}
\usepackage[acronym,indexonlyfirst,nomain]{glossaries}

\setacronymstyle{long-short}
\newacronym{RL}{RL}{reinforcement learning}
\newacronym{HRL}{HRL}{hierarchical reinforcement learning}
\newacronym{IL}{IL}{imitation learning}
\newacronym{BC}{BC}{Behavioral Cloning}
\newacronym{DQN}{DQN}{deep Q-Network}
\newacronym{LfD}{LfD}{Learning from Demonstration}
\newacronym{MR}{MR}{Montezuma's Revenge}
\newacronym{AUC}{AUC}{area-under-the-curve}
\newacronym{ROC}{ROC}{Receiver Operator Characteristic}
\newacronym{VR}{VR}{Virtual Reality}
\newacronym{NN}{NN}{Neural Network}
\newacronym{CNN}{CNN}{Convolutional Neural Network}
\newacronym{ALE}{ALE}{Arcade-Learning-Environment}
\newacronym{iou}{IoU}{Intersection-over-Union}
\newacronym{SVM}{SVM}{Support Vector Machine}
\newacronym{ADAS}{ADAS}{Advanced Driving Assistance Systems}
\newacronym{NMS}{NMS}{non-maximum supression}

\setcopyright{none}
\acmConference[ALA '23]{Proc.\@ of the Adaptive and Learning Agents Workshop (ALA 2023)}
{May 29-30, 2023}{London, UK, \url{https://alaworkshop2023.github.io/}}{Cruz, Hayes, Wang, Yates (eds.)}
\copyrightyear{2023}
\acmYear{2023}
\acmDOI{}
\acmPrice{}
\acmISBN{}
\title[\methodName]{\methodName: Towards \underline{Int}ention-based\\ \underline{H}ierarchical \underline{R}einforcement \underline{L}earning}

\author{Anna Penzkofer}
\affiliation{
  \institution{University of Stuttgart}
  \city{Stuttgart}
  \country{Germany}}
\email{anna.penzkofer@vis.uni-stuttgart.de}

\author{Simon Schaefer}
\affiliation{
  \institution{Technical University of Munich}
  \city{Munich}
  \country{Germany}}
\email{simon.k.schaefer@tum.de}

\author{Florian Strohm}
\affiliation{
  \institution{University of Stuttgart}
  \city{Stuttgart}
  \country{Germany}}
\email{florian.strohm@vis.uni-stuttgart.de}

\author{Mihai B\^ace}
\affiliation{
  \institution{University of Stuttgart}
  \city{Stuttgart}
  \country{Germany}}
\email{mihai.bace@vis.uni-stuttgart.de}

\author{Stefan Leutenegger}
\affiliation{
  \institution{Technical University of Munich}
  \city{Munich}
  \country{Germany}}
\email{stefan.leutenegger@tum.de}

\author{Andreas Bulling}
\affiliation{
  \institution{University of Stuttgart}
  \city{Stuttgart}
  \country{Germany}}
\email{andreas.bulling@vis.uni-stuttgart.de}

\begin{teaserfigure}
    \centering
    \includegraphics[width=0.87\columnwidth]{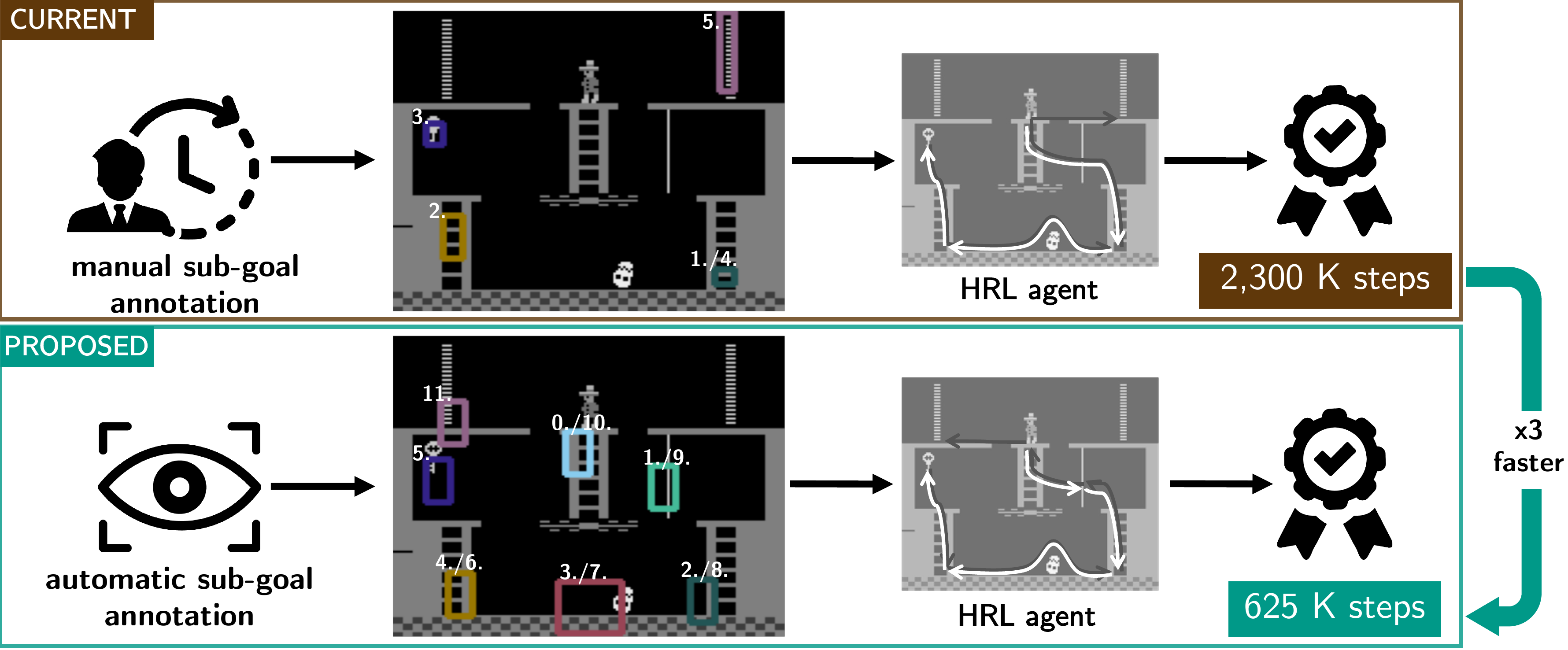}
    \caption{Previous methods require experts to manually annotate meaningful sub-goals for \gls{HRL} agents. 
    We propose automatic sub-goal extraction from human eye gaze, yielding a more robust and sample-efficient \gls{HRL} agent that solves the first room of Montezuma's Revenge from the Atari2600 benchmark after only 625K steps.}
    \label{fig:teaser}
\end{teaserfigure}

\newcommand{\BibTeX}{\rm B\kern-.05em{\sc i\kern-.025em b}\kern-.08em\TeX}
\newcommand{\methodName}{Int-HRL}

\begin{abstract}
While deep reinforcement learning (RL) agents outperform humans on an increasing number of tasks, training them requires data equivalent to decades of human gameplay.
Recent hierarchical RL methods have increased sample efficiency by incorporating information inherent to the structure of the decision problem but at the cost of having to discover or use human-annotated sub-goals that guide the learning process. We show that intentions of human players, i.e. the precursor of goal-oriented decisions, can be robustly predicted from eye gaze even for the long-horizon sparse rewards task of Montezuma's Revenge -- one of the most challenging RL tasks in the Atari2600 game suite.
We propose \textit{\methodName}: Hierarchical RL with intention-based sub-goals that are inferred from human eye gaze.
Our novel sub-goal extraction pipeline is fully automatic and replaces the need for manual sub-goal annotation by human experts.
Our evaluations show that replacing hand-crafted sub-goals with automatically extracted intentions leads to a HRL agent that is significantly more sample efficient than previous methods.  
\end{abstract}

\keywords{Hierarchical Reinforcement Learning, Intention Prediction, Eye Gaze, Montezuma’s Revenge, Sub-goal Extraction}

\begin{document}


\pagestyle{fancy}
\fancyhead{}


\maketitle 


\section{Introduction}
Recent advances in artificial intelligence (AI) in general, and reinforcement learning (RL) in particular, have shown promising results in developing agents that can interact in complex environments and solve challenging real-world tasks, such as robotic manipulation at scale \cite{kalashnkov_mtopt_2021}.
Despite these promising results, a key limitation of RL agents is that training them requires extensive exploration and training data. 
A large body of research~\cite{fan_review_2022, badia_agent57_2020, revaud_r2d2_2019, vinyals2019grandmaster, bellemare_arcade_2013} has thus relied on computer games and other simulated environments to develop and evaluate novel AI agents. 
One of the most popular testbeds are games from the Atari2600 suite implemented in the \gls{ALE}~\cite{bellemare_arcade_2013}.
The Atari2600 games are particularly useful to evaluate RL agents \cite{zhang_atari-head_2020} as they not only have complex visuals but are also challenging for human players~\cite{Mnih2013PlayingAW}. 

Research on the Atari2600 benchmark has focused on deep RL \cite{badia_agent57_2020, badia_never_2020, revaud_r2d2_2019}. While deep RL agents, such as Agent57~\cite{badia_agent57_2020}, have successfully beaten the human benchmark on all 57 Atari games, they are \textit{sample inefficient} and, therefore, require an excessive amount of training. 
Moreover, deep RL methods suffer from a lack of explainability inherent to the deep neural networks used for Q-value estimation.
A more promising approach is hierarchical RL (HRL)~\cite{vezhnevets_feudal_2017,kulkarni_hierarchical_2016, le_hierarchical_2018} that decomposes an RL problem into multiple sub-problems, thus also improving explainability.
A key challenge with HRL is the decomposition of the task that often requires \textit{manual and expert annotations}, which is tedious, time-consuming, and does not easily generalise to other tasks or games.

To address these limitations we propose a novel approach to automatically identify sub-goals in HRL from human eye gaze behaviour. Eye gaze is particularly promising as the gaze location has been linked to human intentions and goals~\cite{huang_using_2015, singh_combining_2020, david-john_towards_2021, chen_gaze-based_2022, belardinelli_gaze-based_2023}.
We hypothesise that these intentions and goals can be further linked to sub-goals so, by predicting players' intentions from their gaze, sub-goals can be identified automatically.
Inspired by prior work on gaze-based intention prediction \cite{huang_using_2015, david-john_towards_2021, chen_gaze-based_2022}, we extract four gaze features and train a \gls{SVM} model. We evaluated the \gls{SVM} on \gls{MR}, a long-horizon sparse reward game from the Atari2600 benchmark, with data from Atari-HEAD \cite{zhang_atari-head_2020}, a data set that offers gaze data in addition to human gameplay demonstrations. Our intention prediction model achieves an average accuracy of 75\%, demonstrating the relation between intention and gaze behaviour, which motivates the automatic extraction of sub-goals for \gls{HRL} agents. Finding useful sub-goals, which is also known as the \textit{option discovery problem} \cite{botvinick_model-based_2014}, is a major issue in \gls{HRL}. However, by using user intents and gameplay demonstrations, our method is able to not only refine and extract the sub-goals, but also the sequence in which these have to be solved to complete the game level. We then integrate the predicted sub-goals into the HRL framework hg-DAgger/Q \cite{le_hierarchical_2018} and show that this approach can solve the first room of \gls{MR} three times more efficiently -- improving sample efficiency from around 2.3 million to around 625 thousand training steps. 

In summary, our work makes two distinct contributions:
(1) We propose a novel method to predict sub-goals for HRL from eye gaze and human demonstration data.
Gaze information is used to predict user intentions that are linked to the sub-goal locations, while demonstration data provides the order in which these sub-goals have to be solved to complete a task.
(2) We evaluate our approach on Montezuma's Revenge from the Atari2600 benchmark and demonstrate significant improvements on two key limitations: sample efficiency and the need for manual expert annotations.
These first results are promising and point towards new intention-based HRL methods that leverage both hierarchical methods and additional human behavioural data, such as eye gaze, to train more efficient agents that can solve complex visual problems. 

\section{Related Work}
\textbf{Hierarchical Reinforcement Learning}.
Deep \gls{RL} has shown great results on the Atari benchmark but still struggles to learn robust value functions from sparse feedback in long-horizon games such as \gls{MR}. Specifically, current state-of-the-art methods require frame samples in the range of billions, which forces researchers to develop elaborate distributed training schemes \cite{revaud_r2d2_2019, badia_agent57_2020} that still take a considerable amount of time to train \cite{fan_review_2022}.
\gls{HRL}, on the other hand, offers a way of exploiting the hierarchical structure of decision-making tasks, guiding the agent towards meaningful sub-goals, effectively increasing the sample efficiency of agents. Moreover, agents achieving consecutive sub-goals, are directly understandable, making \gls{HRL} particularly useful in domains, where explainability is required.

Early on, even before deep \gls{RL}, two ideas have emerged in \gls{HRL}: the options framework \cite{sutton_intra-option_1998} and feudal networks \cite{dayan_feudal_1992}.
Sutton et al. \cite{sutton_intra-option_1998} have proposed to temporally extend actions into \textit{options}, which are composed of a policy, a termination condition, and a set of states in which they could be applied \cite{sutton_intra-option_1998}.
They have shown that Q-learning could be generalised to learning policies over options and that learning inside these options, called "intra-option" learning, allowed the agent to learn about the respective options without executing them explicitly.
Feudal networks, on the other hand, define a hierarchical structure of managers and sub-managers that are only privy to the space and temporal state at their granularity, effectively hiding information from their superior and providing rewards to their sub-managers even if their superior goal was not satisfied \cite{dayan_feudal_1992}. Both hierarchical frameworks have demonstrated much faster convergence than non-hierarchical methods in their respective maze scenarios.

More recently, Kulkarni et al. have proposed a hierarchical approach to induce goal-directed behaviour that does not use separate Q-functions as in the options framework~\cite{kulkarni_hierarchical_2016}.
This made their method scalable and promoted shared learning between options.
To this end, they proposed a two-level framework in which the top-level agent (meta-controller) was responsible for choosing sub-goals while the low-level agent was concerned with achieving these goals. Le et al. have extended this approach by integrating the interactive imitation learning approach DAgger \cite{ross_reduction_2011} into the meta-controller ~\cite{le_hierarchical_2018}. This, however, introduced the need for an expert during training.
Their approach is also similar to feudal networks \cite{vezhnevets_feudal_2017, dayan_feudal_1992} in their hierarchical structure, however, needs significantly less data as it does not use standard \gls{RL} on the higher level.
Vezhnevets et al. have later argued that a disadvantage of\cite{kulkarni_hierarchical_2016} is the need for pre-defined sub-goals and have chosen to learn goal embeddings implicitly \cite{vezhnevets_feudal_2017}.
In this work, we take the best of both worlds and leverage the information provided by gaze data to extract sub-goals independently. This allows us to use the more sample-efficient hierarchically guided method (hg-DAgger/Q) \cite{le_hierarchical_2018}.   

Another work developed concurrently with ours is based on the options framework but also defines intentions as fully satisfied if a sub-goal is reached and evaluates a reduction in available actions to the ones that are affordable in a given state (affordances) via attention. Nica et al. \cite{nica_paradox_2022} introduce these \textit{affordance-aware sub-goal options} with a respective model-free \gls{RL} algorithm and find empirically in a MiniGrid domain that this yields better sample efficiency and higher performance in long-horizon tasks. While they also incorporate visual attention, they do so by applying it to limit an agent's action choices. 
In our work, on the other hand, we use visual attention maps generated from eye gaze data to extract meaningful sub-goals that can be directly selected by the meta-controller of our more feudal network-like architecture. 

\begin{figure*}[t]
  \includegraphics[width=0.87\textwidth,height=3.5cm]{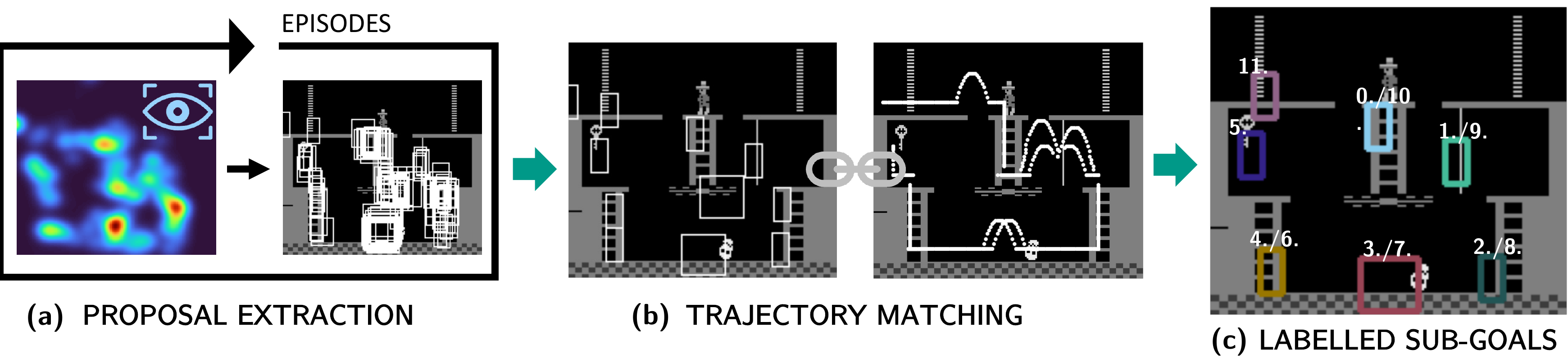}
  \caption{Sub-goal extraction pipeline: \textbf{(a)} proposal extraction is performed from human attention maps for each episode and resulting proposals are merged via \gls{NMS}, then final proposals for one room are matched with human agents' trajectories \textbf{(b)}, yielding labeled sub-goals and visitation order \textbf{(c)}.}
  \label{fig:pipeline}
\end{figure*}

\textbf{Intention Prediction}.
Intentions are goals and desires associated with a concrete plan, i.e. an intention causes a sequence of actions that lead to achieving a certain goal \cite{astington_childs_1994}. In other words, intentions are the precursor of actions, which poses the question whether human intentions are able to pose as sub-goals for \gls{HRL} agents, where the hardest problem is to discover suitable sub-goals \cite{botvinick_model-based_2014}. However, human intention prediction has never been done in this context before. Therefore, before we can replace hand-crafted sub-goals with human intention, we verify whether intention prediction is feasible on \gls{MR}. 

Model-free intention prediction models rely on eye gaze as the most important feature \cite{huang_using_2015, singh_combining_2020, david-john_towards_2021, chen_gaze-based_2022, belardinelli_gaze-based_2023}. 
For a tabular summary of intention and activity recognition using eye gaze in \gls{VR}, PC, table-top, and real-world environments we refer the reader to Chen and Hou \cite{chen_gaze-based_2022}. The work of Huang et al. \cite{huang_using_2015} is the most relevant to ours, as they consider intention prediction as a multi-class classification in a real-world scenario. They achieved 89\% accuracy in their collaborative ingredient prediction task, where a customer instructs a worker to add displayed ingredients to a sandwich, and 76\% accuracy with gaze features alone. The gaze features used in their \gls{SVM} model were: \textit{total duration of looks}, \textit{most recently looked at}, \textit{number of glances}, \textit{duration of first glance}. We successfully test their model on \gls{MR} with gaze data from the demonstration data set Atari-HEAD \cite{zhang_atari-head_2020}. 

Belardinelli \cite{belardinelli_gaze-based_2023} offers a more general review on gaze-based intention estimation, identifying application areas of intention prediction as human-computer interfaces, human-robot interaction, and \gls{ADAS} with relevant works from the last decade of research. However, the application of intentions to solve the \textit{option discovery problem} in \gls{HRL}, or in our case the \textit{sub-goal discovery problem}, is to the best of our knowledge a novel idea and constitutes the main contribution of our work. 


\section{Sub-goal Discovery}
\textbf{Prerequisites}.
\gls{MR} is one of the most challenging games in the Atari2600 suite because of its long planning horizon and sparse rewards \cite{le_hierarchical_2018}. 
A \gls{RL} agent only receives feedback sparingly, requiring many actions to achieve a small reward. Unlike similar long-horizon planning tasks, e.g. artificial grids \cite{nica_paradox_2022, botvinick_model-based_2014}, \gls{MR} is more challenging because it features different rooms that change according to the current level and collecting items allows for different actions in them. Therefore, to identify a specific state of the game it is necessary to know the position of the agent, room ID, level number, and the number of keys held \cite{burda_exploration_2018}. The room ID is particularly important for our method because gaze data should be evaluated separately for each room so that gaze points can be mapped to the specific areas of interest.

The required state information can be extracted directly from the \gls{ALE} via an environment wrapper called Atari Annotated RAM Interface (AtariARI)~\cite{anand_unsupervised_2020}.  
The wrapper parses information from the state variables in the \gls{ALE} and makes it available for each environment step. 
However, the AtariARI wrapper was not used in the collection process of the data set Atari-HEAD \cite{zhang_atari-head_2020}. To acquire the necessary labels subsequently, we simulated the episodes played by humans. This was possible as the original collection was done in a frame-by-frame mode, labeling each consecutive action.

\begin{figure*}[t]
  \includegraphics[width=0.9\textwidth]{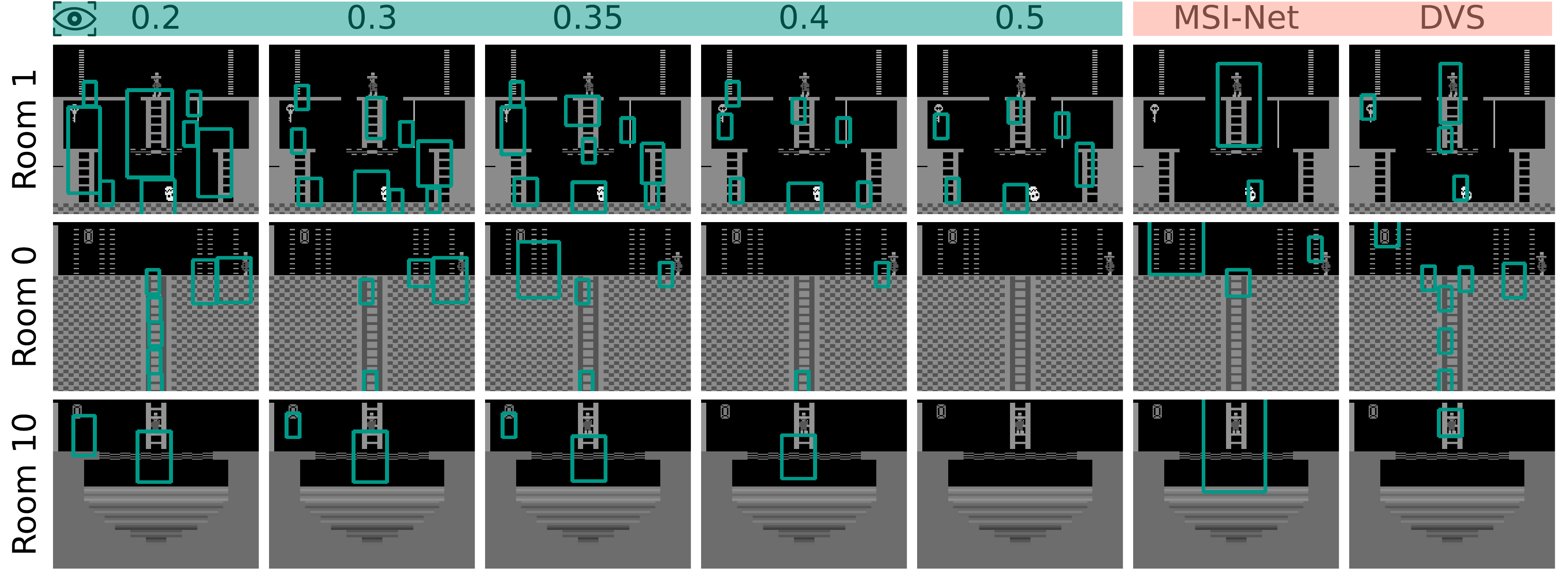}
  \caption{Sub-goal extraction examples on three rooms of \gls{MR} with different saliency map thresholds from $0.2$ to $0.5$ on human gaze data, as well as with saliency maps generated by the MSI-Net \cite{kroner_contextual_2020} and DVS \cite{matzen_data_2018} saliency models with fixed threshold of $0.4$.}
  \label{fig:rooms}
\end{figure*}

\textbf{Sub-goal Extraction}.
Our method for sub-goal extraction is inspired by previous research that showed that visual attention is a predictor of human intentions \cite{huang_using_2015, pfeiffer_eye-tracking-based_2020, david-john_towards_2021, chen_gaze-based_2022, belardinelli_gaze-based_2023} and is further validated by successfully performing intention prediction on the extracted sub-goals in room one. The novel extraction pipeline is visualised in Figure \ref{fig:pipeline}: separate visual saliency maps are calculated for each episode and further isolated to only include the gaze data from the first room in the first level. These saliency maps are generated by adding each gaze point to the frame and passing a Gaussian filter over the generated fixation map with variance $\sigma$ being one visual degree (pixel / visual degrees of the screen). Finally, the saliency map is normalised into the range of $[0, 1]$. An example saliency map can be seen in Figure \ref{fig:pipeline} (a), where hot areas (red/yellow) indicate a high focus of attention across the selected time frame, and cold areas (blue) were not gazed upon at all. 

After generating saliency maps for all episodes, each saliency map was thresholded, i.e. only values above $0.4$ were kept. This threshold was fine-tuned to yield the best results over the entire sub-goal extraction pipeline, a qualitative assessment of other thresholds can be seen in Figure \ref{fig:rooms}. Then, sub-goal proposals were generated, by drawing an agent-sized bounding box around each remaining saliency map point. These proposals were then processed with a custom implementation of the non-maximum suppression algorithm (NMS) \cite{neubeck_efficient_2006}. In general, NMS is applied to suppress overlapping bounding boxes if they exceed an \gls{iou} threshold, and, in our case, boxes with higher saliency values were favoured. Then, the remaining overlapping boxes were merged into one. After the number of sub-goal proposals for each episode was greatly reduced in this manner, the process was repeated to combine proposals across all episodes, yielding a final number of 11 possible sub-goals for room one, as shown in Figure \ref{fig:pipeline} (b).

With the definition of intention in mind, where intention directly leads to goal-directed behaviour \cite{astington_childs_1994}, it is intuitive to only include sub-goals as labels for intentions that are visited during gameplay. Therefore, we ran another simulation of the game data from human players to find the sub-goals that were visited and in which order (trajectory matching). As \gls{MR} is considered to be an almost deterministic game, i.e. there exists a best sequence of sub-goals, this resulted in almost identical orders across episodes. The remaining discrepancies were rectified by implementing a majority vote. 

Overall, this extraction procedure resulted in 7 remaining sub-goals, labeled in order as shown in Figure \ref{fig:pipeline} (c): moving from the middle ladder (0) to rope (1) and bottom right ladder (2), to crossing the middle area with the dangerous and dynamic skull (3), to the bottom left ladder (4), climbing up to collect the key (5), and then reversing this order to get to the left door (11). Interestingly, in all the episodes collected of human gameplay, only the left door was used, most likely because this is the best route suggested in \gls{MR} solution guides.

In comparison to the four hand-crafted sub-goals selected by the \gls{HRL} framework of Le et al. \cite{le_hierarchical_2018} (top row of Figure \ref{fig:teaser}), which they hand-picked from the six sub-goals manually selected by Kulkarni et al. \cite{kulkarni_hierarchical_2016}, our automatic pipeline extracted the same goals and added more areas of interest. In detail, Kulkarni et al. originally performed object detection on the image of room one and then chose the two doors, the three ladders, and the key as entities to define relational goals in the form of \textit{agent} \textit{reaches} \textit{goal}.

\textbf{Sub-goal Analysis}.
We further analysed our sub-goal extraction pipeline (Figure \ref{fig:pipeline}) qualitatively by generating proposals and final sub-goals for additional rooms of \gls{MR}, with different saliency map thresholds, but also with artificial saliency maps generated by saliency models \cite{matzen_data_2018, kroner_contextual_2020}, the results of which can be seen in Figure \ref{fig:rooms}.
We showcase \textit{Room 1} as the starting point of the game, \textit{Room 0} as the second room reached when choosing the left door, and \textit{Room 10} as it features a special room layout. The saliency map threshold is a hyperparameter that needs to be finetuned on the overall extraction, where we have chosen $0.4$ as it includes all hand-crafted sub-goals proposed by prior work \cite{kulkarni_hierarchical_2016, le_hierarchical_2018} with the meaningful addition of the area around the skull, without adding insignificant goals as with lower thresholds, but still including the door, which would be left out by a higher threshold. While there are no hand-crafted sub-goals to compare to for other rooms of \gls{MR}, we can see that the pipeline also selects meaningful sub-goals, e.g. the bottom pathway in \textit{Room 0}, the disappearing floor in \textit{Room 10}, or the diamonds that give an external reward in both. 
In contrast, artificially generated saliency maps by MSI-Net \cite{kroner_contextual_2020}, a standard saliency model with state-of-the-art results on the saliency benchmark CAT2000 \cite{borji_cat2000_2015}, and DVS \cite{matzen_data_2018}, a saliency model optimised for data visualisations, have a predominant focus on the agent itself and otherwise fail to find important steps like the doors, even though they are highlighted in the same colour as the key. Note here that saliency map prediction was done on the RGB images.

\textbf{Intention Prediction}.
For testing the intention prediction model of Huang et al. \cite{huang_using_2015} on our extracted sub-goals, we preprocessed the gaze data following prior work~\cite{david-john_towards_2021, chen_gaze-based_2022}, extracting saccade and fixation events and calculating the four features: \textit{total duration of looks}, \textit{most recently looked at}, \textit{number of glances}, \textit{duration of first glance}.  
We then implemented intention prediction as a multi-class classification for the 7 sub-goals of room one with a \gls{SVM}. We achieve an average prediction accuracy of 75\% in a 10-fold cross-validation, which is significantly better than a random model and also outperforms results reported on other data sets \cite{huang_using_2015, belardinelli_gaze-based_2023}. We argue that this corroborates the efficacy of using human gaze data as an indicator of intention and motivates the extraction of sub-goals for \gls{HRL} from human intention. 

\section{Intention-based Learning}
\textbf{Baseline}.
One approach for solving long-horizon decision-making tasks is \gls{HRL}, where two popular frameworks emerged in the past: the options framework \cite{sutton_intra-option_1998, nica_paradox_2022} and feudal networks \cite{dayan_feudal_1992, vezhnevets_feudal_2017}. Building upon a feudal architecture, by combining deep \gls{HRL} with pre-defined sub-goals, Kulkarni et al. \cite{kulkarni_hierarchical_2016} are able to outperform na\"ive deep Q-learning. 
Their h-DQN model was tested on two delayed-reward domains, including the first room of \gls{MR}, where their approach is able to reach the door after 2.5\,M samples. Taking the idea further, Le et al. \cite{le_hierarchical_2018} combined \gls{IL} with \gls{HRL} showing that their hierarchical guidance model (hg-DAgger/Q) significantly reduces expert effort compared to other interactive \gls{IL} approaches and is also able to learn faster and more robustly than h-DQN, solving the first room of \gls{MR} after 2.3\ M steps. 

Our baseline, the hg-DAgger/Q model by Le et al., consists of two levels, the meta-controller level, and the agents level. They use the data aggregation method DAgger \cite{ross_reduction_2011} on the top level, which trains the meta-controller policy with iteratively aggregated data sets. The meta-controller is used to predict one of the four hand-crafted sub-goals, initiating the corresponding agent. 

On the low level, Le et al. used a double deep Q-Network (DDQN) \cite{van_hasselt_deep_2015} with prioritised experience replay \cite{schaul_prioritized_2016}. A major difference between their approach and h-DQN \cite{kulkarni_hierarchical_2016} is that separate agents are trained for each sub-goal instead of passing the goal vector as a feature into a single policy network. This ensures the mitigation of the issue of catastrophic forgetting and also has the advantage of separate exploration schedules. However, maintaining a separate network for each sub-goal is not scalable across different rooms of \gls{MR}. 

Combining the low-level \gls{RL} agents with hierarchical guidance from the meta-controller ensures that the experience buffer for the DDQN only contains valuable samples for the next sub-goal, as wrong meta-controller choices terminate the episode. Le et al. argue that this is the main reason for their higher robustness in training.
However, Le et al. also report that their architecture only learned all sub-goals successfully in 50 out of 100 trials. This high variability is most likely due to different implementations and random seeding, an issue common in \gls{RL} \cite{henderson_deep_2019}, which would also explain, why we were unable to reproduce their results. Consequently, we will compare our results to the ones reported in their paper. 

In summary, hg-DAgger/Q \cite{le_hierarchical_2018} is significantly more sample efficient than other methods \cite{vezhnevets_feudal_2017, kulkarni_hierarchical_2016}. However, it requires an expert at training time to select hand-crafted sub-goals and only implements a rudimentary sub-goal check.

\textbf{Model}.
Similar to \cite{le_hierarchical_2018}, we use a hierarchical reinforcement learning approach, with $8$ possible actions (no action, cardinal moving directions, jumping up, left, and right). 
Starting with a custom implementation of hg-DAgger/Q \cite{le_hierarchical_2018}, we tested different approaches to making training more stable. By using a Dueling \gls{DQN} architecture \cite{wang_dueling_2016} in addition to the DDQN \cite{van_hasselt_deep_2015} and including the lower left ladder (see Fig. \ref{fig:teaser} top row goal 2) as an additional hand-crafted goal, we were able to train a model to reach the first external reward, the key. 

To further improve the model performance, we replace the hand-crafted sub-goals with the fields of interest derived from human gaze data, as previously described, thereby expanding the set of sub-goals used in \cite{le_hierarchical_2018, kulkarni_hierarchical_2016}. 
Sub-goals are a way of providing pseudo rewards \cite{le_hierarchical_2018, kulkarni_hierarchical_2016, sutton_intra-option_1998} to populate the sparse reward map in \gls{MR}. 
Next to the sub-goal reached reward $R_{\text{sub-goal}}$, we introduce a dense reward signal $R_{\text{dir}}$,  $R_{\text{dist}}$, and $R_{\text{step}}$ to further stabilise training.

\begin{align}
    R = R_{\text{sub-goal}} + \alpha R_{\text{dir}} + \beta R_{\text{dist}} + \gamma R_{\text{step}}
\end{align}

We define a direction reward $R_{\text{dir}}$ to steer the agent in the direction of the next sub-goal. It is computed as the scalar product of the selected action's direction vector $\vec{a}$ and the vector between the next and previous goal $\vec{g} = G_{prev} - G_{next}$: 

\begin{align}
    R_{\text{dir}} &= <\vec{a}, \vec{g}>
\end{align}

The distance reward $R_{\text{dist}}$ guides the agent to the next sub-goal by minimising the euclidean distance between the agent and current goal $d_{ac}$, as well as previous goal $d_{ap}$, and the distance between previous and current goal $d_{pc}$:

\begin{align}
    R_{\text{dist}} &= \frac{\sqrt{d_{ap}} - \sqrt{d_{ac}}}{\sqrt{d_{pc}}}
\end{align}

The step reward $R_{\text{step}}$ penalises each time-step used to reach the next goal with a constant $\tau = 0.001$, which is also used to scale $R_{\text{dist}}$ and $R_{\text{dir}}$.

The distance reward $R_\text{dist}$ requires the agent location at each step, we propose to either use a specifically trained object detector, which was tested with a pre-trained FasterRCNN model \cite{ren_faster_2016} fine-tuned on ~100 manually labelled training examples, or to use the RAM state labels provided via the AtariARI Wrapper \cite{anand_unsupervised_2020}.

Both of these approaches are much more robust than the rudimentary approach used in \cite{le_hierarchical_2018} and, additionally, are able to track non-static regions of interest. 

\begin{figure*}[t]
  \includegraphics[width=0.8\textwidth]{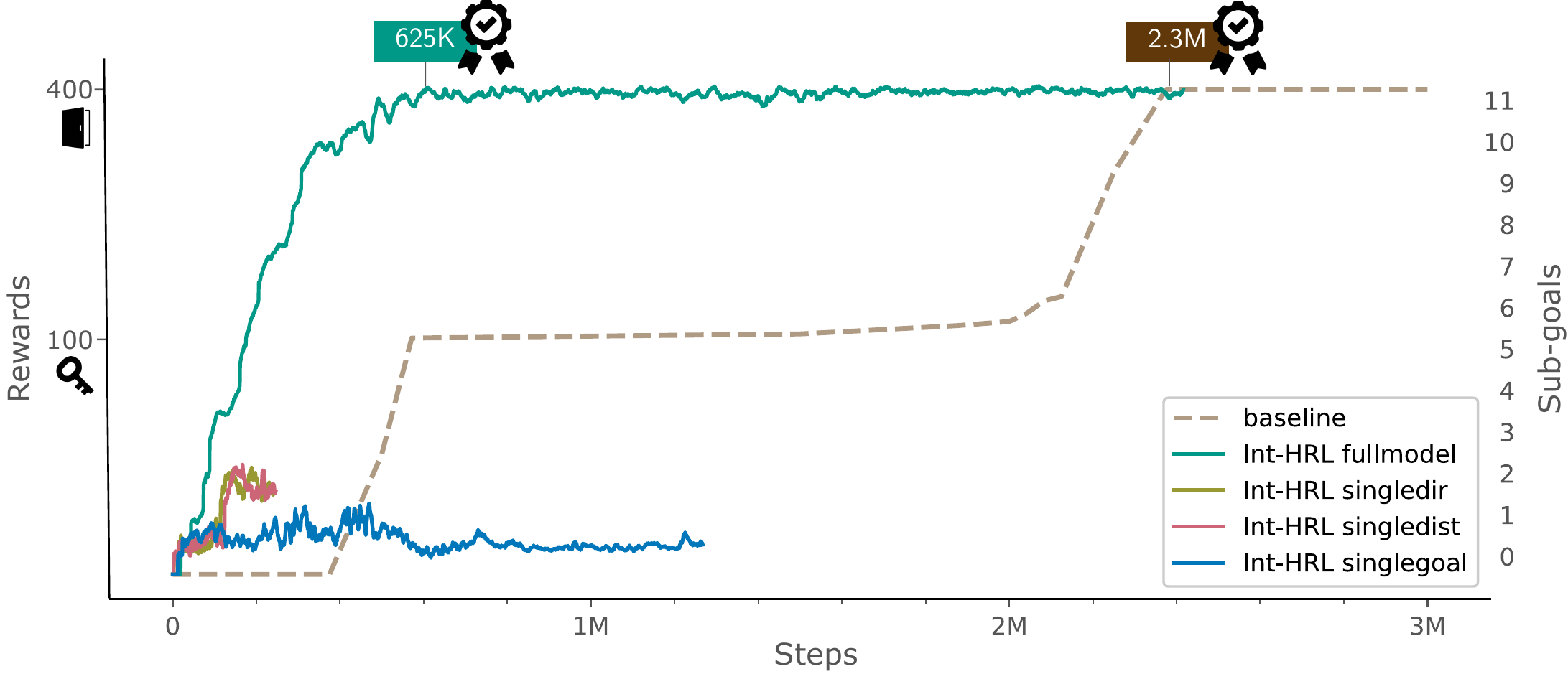}
  \caption{Step-wise sub-goal trailing performance of \methodName and baseline with sub-goal 5 as a first external reward from the key and sub-goal 11 the right/left door, which completes the first room.}
  \label{fig:results}
\end{figure*}

\textbf{Results}. We evaluated the full-sequence hierarchical model without any dense reward (\textit{fullmodel}), a single-agent model with negative step reward and goal feature: $\gamma = 1$ (\textit{singlegoal}), and the single-agent model with a distance reward: $\beta = 1$ (\textit{singledist}) in comparison to the direction reward: $\alpha = 1$ (\textit{singledir}). 
To further help the single-agent models, we passed the sub-goals as additional feature vectors.
Throughout all experiments, we use an $\epsilon$-greedy exploration policy with a linear scheduler ranging from $1$ to $0.02$ in $200$\ K steps, a prioritized replay buffer \cite{schaul_prioritized_2016}, and a  learning rate of $0.0001$.
Results are shown in Figure \ref{fig:results}. Each sub-goal is considered learned when the trailing performance is above 0.9, i.e. when the agent reaches the goal 90 out of 100 trials. 
Directly compared models are trained with the same random seed to ensure comparability~\cite{henderson_deep_2019}.

The full-sequence model is the first to successfully learn to solve the first room of \gls{MR} by reaching the left door. 
The final goal (11) is consistently reached by the \textit{fullmodel} after only 625K steps. 
Hence, our model is more than three times more sample efficient than our baseline \cite{le_hierarchical_2018}, which needs 2.3M samples to complete the first room. Furthermore, our new framework works without an expert, as we extracted the chosen sub-goals previously from gaze data and perform a true goal reached check via RAM state mapping. Overall, given labelled gaze demonstration data, the pipeline can be trained end-to-end with no manual effort.

In comparison to the full-sequence model, the single-agent model with the goal feature and small negative rewards per step does not succeed at all. As expected, the single Dueling DDQN agent suffers from a lack of \textit{exploration} \cite{le_hierarchical_2018}, i.e. as Figure \ref{fig:results} shows, the \textit{singlegoal} model fails to learn anything new and gets stuck at the first sub-goal. Resetting the schedule when a new sub-goal is explored was tested, but did not succeed as it resulted in $\epsilon$ being too high which also prevents learning.
Further ideas for solving insufficient exploration are included in the discussion and are left for future work.  

Providing a more dense reward structure by adding a distance or direction reward improves the single-agent model as it increases sample efficiency so that the model learns to reach sub-goal 4, i.e. passing the skull and reaching the lower left ladder (see the \textit{singledist} and \textit{singledir} model in Figure \ref{fig:results}). In our trial, the direction reward even facilitates learning to reach the key sub-goal (5), which provides the first external reward. Interestingly, the direction reward is simpler to implement as it does not require knowledge of the agent's location. Both models still suffer from the issues encountered by the single-agent model approach and performance deteriorates after 200K steps when the $\epsilon$-exploration schedule reaches its final value. However, this confirms that more intrinsic rewards improve performance and should therefore be incorporated into future models.

\section{Discussion and Outlook}
We have shown that gaze features are indicative of intentions by successfully training a simple intention prediction model on the first room of the Atari game MR. The model predicts the next intention of an agent with 75\% accuracy, thus validating the relation between eye gaze and intentions, and could be easily extended to a more powerful classification model by using more detailed fixation and saccade dynamics \cite{david-john_towards_2021}, or with neural networks and temporal information \cite{belardinelli_gaze-based_2023}. 
Further, we developed a novel sub-goal extraction pipeline from gaze data. To this end, we labelled an available demonstration data set via simulation, analysed the visual attention heatmaps for each room, and aligned the proposals with the agent trajectories. This process yields sub-goals that are on par with hand-crafted ones from prior work \cite{le_hierarchical_2018, kulkarni_hierarchical_2016}. 
We demonstrate the efficacy of our sub-goal extraction pipeline by using the extracted sub-goals to train an HRL agent that can solve the first room of MR significantly more sample efficiently than any previous method. Moreover, our pipeline is fully automatic and allows for a transparent explanation of agent behaviour. In comparison to previous methods, where sub-goals have been chosen manually without further analysis \cite{le_hierarchical_2018, kulkarni_hierarchical_2016}.

\textbf{Generalisability to other games.} One requirement for our approach is a fixed layout of rooms for extracting meaningful information from gaze data. 
Areas of interest need to be stationary enough for a high duration of attention and depict isolated or unique objects.
We have chosen the long-horizon sparse reward game \gls{MR} of the Atari2600 suite because it can be structured into sub-tasks across different static rooms and standard \gls{RL} still struggles with solving it efficiently. Other games available in the Atari-HEAD \cite{zhang_atari-head_2020} data set are not suitable for this analysis because agents and sprites can move across all lanes, because objects of interest scroll too fast across the game screen, or because areas of interest are trivial, as in shooter games where all sprites are at the top of the screen and agent movement is restricted to be horizontal across the bottom. 
Other games similar to \gls{MR} are \textit{Hero} and \textit{Venture}, where different rooms need to be navigated, which include static sprites or objects and clear goals. While \textit{Hero} is structured like a search tree, iteratively expanding the depth of exploration for solving a level by finding people lost in the caverns, \textit{Venture} is like a Maze with an overview screen from which the agent can reach different rooms to find treasure. They were not selected because they are not considered particularly difficult for \gls{RL}; however, it will be interesting to see whether the intention predictor can add explainability to agents for these games in future work. 

\textbf{Sacalability of \gls{HRL} method.} While the full-sequence model has outperformed all other baselines tested on the first room of \gls{MR} in terms of sample efficiency \cite{kulkarni_hierarchical_2016, vezhnevets_feudal_2017, le_hierarchical_2018}, it needs to be more scalable to solve the entire game. The \gls{HRL} approach based on Le et al. \cite{le_hierarchical_2018} requires the separate handling of 12 agents in the first room of \gls{MR} but an additional 23 rooms need to be explored to solve the first level. While both \cite{kulkarni_hierarchical_2016, vezhnevets_feudal_2017} only use a single low-level agent, the former's successful trials were not reproducible \cite{le_hierarchical_2018} and the latter requires 200M samples for the first room alone, most likely getting close to the 10 billion samples needed by standard deep \gls{RL} approaches \cite{badia_agent57_2020}. We have tested single agents with more dense reward structures ($R_\text{step}$, $R_\text{dist}$, $R_\text{dir}$), however, were unable to circumvent the issue of \textit{insufficient exploration} in single agents. In future work, we would like to address this by adding per-episode and full-game novelty values as intrinsic rewards, which succeeded in deep \gls{RL} methods \cite{badia_agent57_2020, badia_never_2020}.

\balance



\begin{acks}

Anna Penzkofer was funded by the Deutsche Forschungsgemeinschaft (DFG, German Research Foundation) under Germany’s Excellence Strategy -- EXC 2075 -- 390740016.
Simon Schaefer was supported by TUM AGENDA 2030, funded by the Federal Ministry of Education and Research (BMBF) and the Free State of Bavaria under the Excellence Strategy of the Federal Government and the Länder as well as by the Hightech Agenda Bavaria.
Mihai Bâce was funded by a Swiss National Science Foundation (SNSF) Postdoc.Mobility Fellowship (grant number 214434).
Florian Strohm and Andreas Bulling were funded by the European Research Council (ERC) under the grant agreement 801708.

\end{acks}



\bibliographystyle{ACM-Reference-Format} 
\bibliography{bibliography}

\end{document}